\documentclass[journal]{IEEEtran}
\usepackage[tbtags]{amsmath}

\usepackage{multirow}
\usepackage{color}
\usepackage{threeparttable}
\newcommand{\tabincell}[2]{\begin{tabular}{@{}#1@{}}#2\end{tabular}}

\usepackage{cite}

\ifCLASSINFOpdf
  \usepackage[pdftex]{graphicx}
  \graphicspath{{./pictures/}}
\else
\fi
\usepackage{amsmath}
\usepackage{algorithm}
\usepackage{algpseudocode}

\usepackage{array}
\usepackage{booktabs}
\begin{document}
%
\title{Auto Machine Learning for Medical Image Analysis\\ by Unifying the Search on Data Augmentation and Neural Architecture}

%

\author{
	Jianwei~Zhang,
	Dong~Li,
	Lituan~Wang,
	Lei~Zhang,~\IEEEmembership{Senior Member,~IEEE,}

	\thanks{This work was supported by Natural Science Foundation of China under Grant No.62025601, 61772353.~\emph{(Corresponding author: Lei Zhang.)}}
	
	\thanks{The authors are with the Machine Intelligence Laboratory, College of
	Computer Science, Sichuan University, Chengdu 610065, China (e-mail: zhangjianwei@stu.scu.edu.cn; lidong@stu.scu.edu.cn; lituanwang@scu.edu.cn; leizhang@scu.edu.cn).}
}

\IEEEtitleabstractindextext{%
\begin{abstract}
	Automated data augmentation, which aims at engineering augmentation policy automatically, recently draw a growing research interest. Many previous auto-augmentation methods utilized a Density Matching strategy by evaluating policies in terms of the test-time augmentation performance. In this paper, we theoretically and empirically demonstrated the inconsistency between the train and validation set of small-scale medical image datasets, referred to as in-domain sampling bias. Next, we demonstrated that the in-domain sampling bias might cause the inefficiency of Density Matching. To address the problem, an improved augmentation search strategy, named Augmented Density Matching, was proposed by randomly sampling policies from a prior distribution for training. Moreover, an efficient automatical machine learning(AutoML) algorithm was proposed by unifying the search on data augmentation and neural architecture. Experimental results indicated that the proposed methods outperformed state-of-the-art approaches on MedMNIST, a pioneering benchmark designed for AutoML in medical image analysis.
	
	
\end{abstract}

\begin{IEEEkeywords}
Automated data augmentation, Medical image analysis, Auto machine learning, Neural architecture search.
\end{IEEEkeywords}
}

\maketitle

\IEEEdisplaynontitleabstractindextext

%
\IEEEpeerreviewmaketitle

\section{Introduction}
%
%
%
%

\IEEEPARstart{I}{n} recent years, deep learning, a subfield of machine learning (ML), has become the mainstream method in computer-aided diagnosis (CAD) and medical image analysis\cite{esteva2019guide, budd2021survey, xie2021survey, wang2020automatic}. Due to its powerful ability in computer vision (CV), many studies have shown state-of-the-art performance in many key areas of medicine, such as radiology\cite{yu2020multitask, wang2021deep, qi2020automated}, dermatology\cite{li2020transformation}, and pathology\cite{cirecsan2013mitosis}.  However, constructing an efficient deep learning application for medical image analysis usually suffers from substantial human labor and enormous time. Automated machine learning (AutoML), which aims for automated construction of the machine learning pipeline under a limited computational budget, recently has drawn a growing research interest\cite{2021AutoML}.

AutoML refers to almost all components of the machine learning pipeline, in which neural architecture is the most crucial part due to its importance on feature extraction of data\cite{ren2020comprehensive}. Hence, recent AutoML studies mainly focus on neural architecture search (NAS), which aims at finding optimal neural architectures across an ample search space within a limited computational budget. 

In aspect of search strategy, NAS approaches could be mainly divided into three categories: reinforcement learning (RL) based methods\cite{zoph2016neural, pham2018efficient, zoph2018learning}, evolutionary algorithm (EA) based methods\cite{real2017large, guo2019single, chu2019fairnas, zhang2021one}, and differential methods\cite{liu2018darts, xie2018snas, cai2018proxylessnas}. The RL-based work\cite{zoph2016neural} could be considered as pioneering work of NAS, in which a recurrent neural network (RNN) is constructed to generate different architectures. The accuracies of the generated architectures on a validation set are utilized to optimize the RNN. For EA-based methods, a population of neural architectures is maintained, where each individual is essentially a neural architecture. The search process is actually a number of evolutionary generations for the population, and then the survived architectures are outputted as the results. Unlike RL-based methods or EA-based methods, which treat the search process as a black-box optimization problem, differential methods formulate the architecture representation continuously so that the search could be directly optimized by gradient descent\cite{ruder2016overview}. With the surge of interest in NAS, a series of studies successfully applied NAS into medical image analysis and achieved state-of-the-art performance\cite{zhu2019v, yu2020c2fnas, guo2020organ}.

Apart from neural architecture, a large amount of data is another crucial factor of deep learning applications' success. However, due to the expensive labor cost, it is always too difficult to collect enough well-labeled data for medical image domain. Data augmentation is an efficient technique to enrich the diversity of training data and reduce the risk of overfitting. Inspired by NAS, automated data augmentation (auto-augmentation), which aims at searching for optimal augmentation strategies automatically, recently draw a significant amount of research efforts\cite{cubuk2019autoaugment, lim2019fast, hataya2020faster, ho2019population, cubuk2020randaugment}. Most of the studies utilized the analogous techniques applied in NAS. AutoAugment\cite{cubuk2019autoaugment} constructed an RNN controller to generate augmentation policies where each policy is constituted by a number of sub-policies. One sub-policy is a sequence of augmentation operations applied to train data. The controller is optimized with RL in terms of the validation accuracy of a fixed neural architecture trained with the generated policy by randomly sampling one sub-policy for each batch's training. The searched augmentation policy yielded state-of-the-art performance on several natural image datasets but suffered from the expensive computation. FastAugment\cite{lim2019fast} utilized an efficient search strategy named Density Matching and Bayesian optimization to search augmentation. As no extra training processes are necessary for policy evaluation, the whole search time could be reduced by orders of magnitude. 

Unfortunately, previous auto-augmentation studies mostly focus on searching optimal augmentation strategies on natural image datasets. However, a significant gap exists between nature images and medical images. For example, medical image datasets always suffer from small data scale and class imbalance (CB). Previous works search optimal augmentation policies based on the validation accuracy, which is calculated through a model trained with the original train set, may not be efficient to guide the search on small-scale medical images. In aspect of NAS, most previous studies search optimal architectures based on cross entropy loss or validation accuracy, which are not appropriate and less sensitive for architecture evaluation on class-imbalanced data. Moreover, most previous AutoML works studied the neural architecture search and data augmentation search separately, which is inefficient and against the AutoML spirit. 

To address the aforementioned problems, we firstly theoretically and empirically demonstrated the inconsistency between the train and validation set of small-scale medical image datasets, which is called \textbf{in-domain sampling bias} in this paper for simplicity. Then, we demonstrated that the in-domain sampling bias might weaken the efficiency of Density Matching and proposed an improved version named \textbf{Augmented Density Matching} to address the problem of vanilla Density Matching. Finally, following the EA-based NAS framework\cite{zhang2021one}, an efficient AutoML algorithm is proposed to unify the search on data augmentation and neural architecture, named \textbf{USAA} for short. Hence, the augmentation policy and neural architecture could be searched in one round of process, dramatically improving AutoML applications' efficiency. During the search process, the Area under the Receiver Operating Characteristics Curve (AUC) is utilized as the main optimization objective to handle the class imbalance problems. The experiments showed that our results outperformed state-of-the-art AutoML algorithms on MedMNIST\cite{yang2021medmnist}, which is a lightweight AutoML benchmark consisting of 10 medical image datasets. 

Our contributions can be summarized as follows:
\begin{enumerate}[]
	\item We theoretically and empirically demonstrated the in-domain sampling bias of small-scale medical image datasets, which may weaken the efficiency of Density Matching. 
	\item To improve the augmentation search efficiency, an efficient augmentation search strategy named Augmented Density Matching is proposed by training with sampled augmentation policies from a distribution.
	\item An efficient AutoML algorithm named USAA is proposed to unify the search on data augmentation and neural architecture. As a result, the optimal neural architecture and augmentation policy could be searched within one round of search process. 
\end{enumerate}

\section{Related Works}\label{related}
In this section, some relevant works are introduced in two aspects: neural architecture search and automated data augmentation.

\subsection{Neural Architecture Search}\label{related_nas}
Due to its high efficiency in engineering networks automatically, neural architecture search could be considered as one of the most crucial sub-domain of AutoML. Early NAS methods are usually designed by treating the architecture search as a black-box optimization problem, which is logically optimized by reinforcement learning or evolutionary algorithm. 
For both the above two categories of methods, a large number of networks have to be trained from scratch to evaluate their performance, thus thousands of GPU days are required to ensure the search results.  To improve the search efficiency, one-shot NAS methods\cite{pham2018efficient, brock2017smash, bender2018understanding} are proposed by designing a \emph{supernet}, which synthesizes all the possible candidate architectures, to represent the whole search space. Hence, all the candidate architectures could be evaluated by just inheriting the sharing weights of supernet, and only one supernet needs to be trained during the whole search process. 

The one-shot methods could be mainly divided into two categories: differential NAS and single-path NAS. Differential NAS reformulated the architecture representation continuously, which could be directly learned through gradient descent\cite{liu2018darts, xie2018snas, liang2019dartss, xu2019pc, chu2020fair}. 
However, in differential NAS, the neural operations with greater weights in the early search stage always have more chance to converge, which inevitably leads to the rich-get-richer problem\cite{adam2019understanding} and local optima\cite{xie2020weight}. For single-path NAS, the supernet training stage and the architecture search stage are decoupled\cite{bender2018understanding, guo2019single}. In the supernet training stage, a path dropout strategy is usually utilized to uniformly train the weights of different neural operations and constrain the sharing weights to evaluate different architectures fairly. In the architecture search stage, different strategies, such as evolutionary algorithm or greedy strategy, could be used for selecting optimal architectures\cite{bender2018understanding}. Moreover, Li et al.\cite{li2020improving} theoretically demonstrated that it was efficient to improve the search efficiency by pruning the search space during the search process. To generalize the pruning strategy to a more complicated search space, such as DARTS\cite{liu2018darts}, Zhang et al.\cite{zhang2021one} proposed a hierarchically-ordered pruning strategy named HOPNAS to guide the search direction. 

For the search space, \cite{liu2018darts}, \cite{xu2019pc}, \cite{zhang2021one} are related to our work, because we all utilize the cell-based search space, which is appropriate to adjust the neural capacity for diverse-scale medical image datasets. For the search strategy, the proposed unified search method belongs to the single-path NAS category. Most related to us is HOPNAS\cite{zhang2021one}, we follow the same NAS framework to search both the data augmentation policy and neural architecture due to its efficiency in individual encoding. What's more, we both prune the search space during the search to improve the search efficiency. Nevertheless, the difference is that, in our method the augmentation operation is treated coessentially as a neural operation. As a result, NAS could be further expanded in a broad sense, which covers the augmentation search. 


\subsection{Automated Data Augmentation}\label{related_autoaug}
With the emerging interest in AutoML, researchers recently began to pay attention to automatically searching augmentation policies, which is essential for data-hungry domains such as medical image analysis\cite{xu2020automatic}. Inspired by NAS, most auto-augmentation algorithms utilized analogous techniques to search optimal augmentation policies. AutoAugment\cite{cubuk2019autoaugment} could be considered as the pioneering work in auto-augmentation. Similar to \cite{zoph2016neural}, an RNN controller is built to generate augmentation policies, which contains all the information about how to augment the train data. Then, the generated policies will be combined with a fixed architecture for training, whose validation accuracy will be utilized as the reward to train the controller through reinforcement learning. However, a large number of policy evaluations usually lead to an expensive search cost. Lim et al.\cite{lim2019fast} proposed an auto-augmentation method called FastAugment based on a Density Matching strategy, which evaluated augmentation policies in terms of the test-time data augmentation performance. Bayesian optimization, which is also widely used in NAS\cite{mendoza2016towards}, is applied to explore optimal augmentation sub-policies.

In contrast to AutoAugment\cite{cubuk2019autoaugment}, FastAugment did not need to repeatedly train a large number of models from scratch for evaluation, thus could significantly reduce the search budget. Similar to SNAS\cite{xie2018snas}, Hataya et al.\cite{hataya2020faster} proposed an algorithm called FasterAugment by relaxing the discrete augmentation distribution via reparameterization trick, enabling augmentation representation parameters to be trained directly by gradient descent. In addition, RandAugment\cite{cubuk2020randaugment} demonstrated that it was sufficient to search optimal augmentation policies across a simplified augmentation search space in which each augmentation operation only has a single distortion magnitude. As a result, only two free parameters, the number of augmentation operations and the magnitude for all operations, are required to be optimized, and grid search is sufficient for searching optimal augmentation policy.

Consistent with the above auto-augmentation studies, we also follow the policy-based augmentation search space, in which the search objective is a diverse and stochastic mix of sequential augmentation operations instead of a fixed one. Nevertheless, the difference is that the search space in the proposed method is implemented by maintaining a population of networks, with different augmentation sub-policies embedded into the individual encoding of EA-based framework\cite{zhang2021one}. In terms of search space, most related to us is RandAugment\cite{cubuk2020randaugment}. We both simplify the search space to improve the efficiency, and the results have also been proved that it is sufficient to search optimal augmentation policies even under the sacrifice of search diversity. The difference is that RandAugment searches a single magnitude for all the augmentation operations, while the proposed Augmented Density Matching abandons the search on augmentation magnitude and applies a small augmentation degree for all the operations with magnitudes to further improve the search efficiency. In terms of metrics for evaluating augmentation sub-policies, FastAugment\cite{lim2019fast} and RandAugment\cite{cubuk2020randaugment} are most related to our work. We all utilize the test-time augmentation performance to evaluate sub-policies, while the difference is that the evaluation in our methods is based on the model trained with augmentation sub-policies sampled from a maintained population. With this novel design, the difference of a model's performance on train set and validation set is shrunk by improving the diversity of train data, which makes it more efficient when handling small-scale medical image datasets. 

\section{Method}\label{method}
In this section, the proposed augmentation search strategy, Augmented Density Matching, is introduced in Subsection~\ref{augmented_density_matching} at first. Then, the search framework which unifies the search on data augmentation and neural architecture is illustrated in Subsection~\ref{usaa}. For convenience, some important notations used in the paper are summarized in Table~\ref{noation_table}.

\renewcommand\tabcolsep{7.0pt} 
\begin{table}
	\centering 
	\caption{Notations}\label{noation_table} 
	\begin{tabular}{@{}ll@{}}
		\toprule
		Symbols & Descriptions \\
		\midrule
		
		$D_{train}$  	    & the train dataset with $N_{train}$ samples\\ 
		$D_{val}$  	    & the validation dataset with $N_{val}$ samples\\ 
		$\mathcal{N}(\mu, {\sigma}^2)$  	& Gaussian distribution with mean $\mu$  and standard deviation $\sigma$\\ 
		$\triangle X$        &  the difference of sample means of $D_{train}$ and $D_{val}$\\ 
		$\mathcal{T}$  & augmentation policy, which has $K$ sub-policies\\ 
		$S^{(k)}$     & augmentation sub-policy, $k=1, 2, ..., K$\\ 
		$\mathcal{O}_a$     &  the set of all candidate augmentation operations\\ 
		$\mathcal{O}_n$     &  the set of all candidate neural operations\\  
		$\theta$   & the network weights\\ 
		$P$  & the population in evolutionary algorithm  \\ 
		\bottomrule
	\end{tabular}
\end{table}

\subsection{Augmented Density Matching}\label{augmented_density_matching}
In this subsection, the inconsistency between train and validation set in small-scale medical image datasets, which is named in-domain sampling bias in this paper for simplicity, is theoretically and empirically demonstrated through the perspective of sampling distribution in Subsubsection~\ref{indomainbias} at first.
Then, the dark side of Density Matching is theoretically demonstrated in Subsubsection~\ref{darkside}. Finally, an improved augmentation search strategy named Augmented Density Matching is proposed to improve the efficiency of vanilla Density Matching in Subsubsection~\ref{improving_density_matching}.

\subsubsection{In-domain Sampling Bias}\label{indomainbias}
Let a train set $D_{train}$ and a validation set $D_{val}$ are independent and identically sampled from a Gaussian distribution $\mathcal{N}(\mu, {\sigma}^2)$, where the number of training and validation samples are $N_{train}$ and $N_{val}$, and the mean of $D_{train}$ and $D_{val}$ are denoted by ${\bar X}_{train}$ and ${\bar X}_{val}$, respectively. For the distribution of sample means, we have:
\begin{gather} \label{eqm1}
{\bar X}_{train} \sim \mathcal{N}(\mu, \frac{{\sigma}^2}{N_{train}} ), \notag \\
{\bar X}_{val} \sim \mathcal{N}(\mu, \frac{{\sigma}^2}{N_{val}} ).   
\end{gather}

Moreover, we could get the property for the difference of sample means:
\begin{equation}\label{eqm2}
\triangle X \sim \mathcal{N}(0, (\frac{1}{N_{train}} + \frac{1}{N_{val}}){\sigma}^2),
\end{equation}
where $\triangle X = {\bar X}_{train} - {\bar X}_{val}$.

\begin{figure}
	\includegraphics[width=0.49\textwidth]{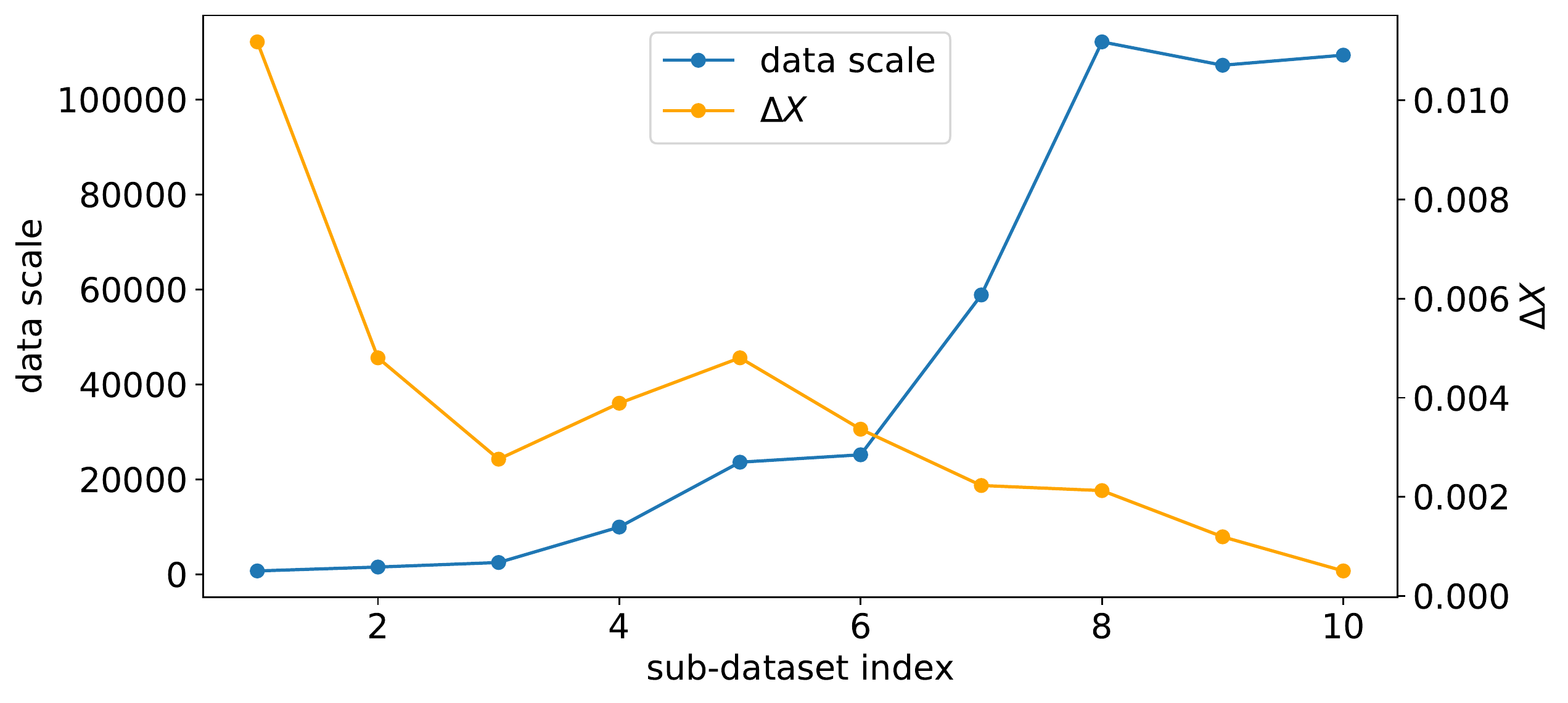}
	\caption{The inverse proportion relation between data scale and the difference of sample means of MedMNIST's 10 sub-datasets. The blue line denotes the data scale, while the orange line denotes the difference of sample means of train set and validation set. There exists a reverse proportion relation between the data scale and the difference.}
	\label{fig_difference}
\end{figure}

According to Eq.~(\ref{eqm2}), the numbers of samples in the train and validation set are smaller, the difference of sample means have more probability to be large. To demonstrate it, the data scale and the difference of sample means are measured on the 10 sub-datasets of MedMNIST. The reverse proportion relation between the data scale and the difference of sample means is shown in Fig.~\ref{fig_difference}. In this paper, we refer the inconsistency between train and validation set in small-scale medical image datasets as in-domain sampling bias, and hypothesize that it could cause the inconsistency of a model's performance  on $D_{train}$ and $D_{val}$.

\begin{figure*}
	\includegraphics[width=1\textwidth, height=0.4\textwidth]{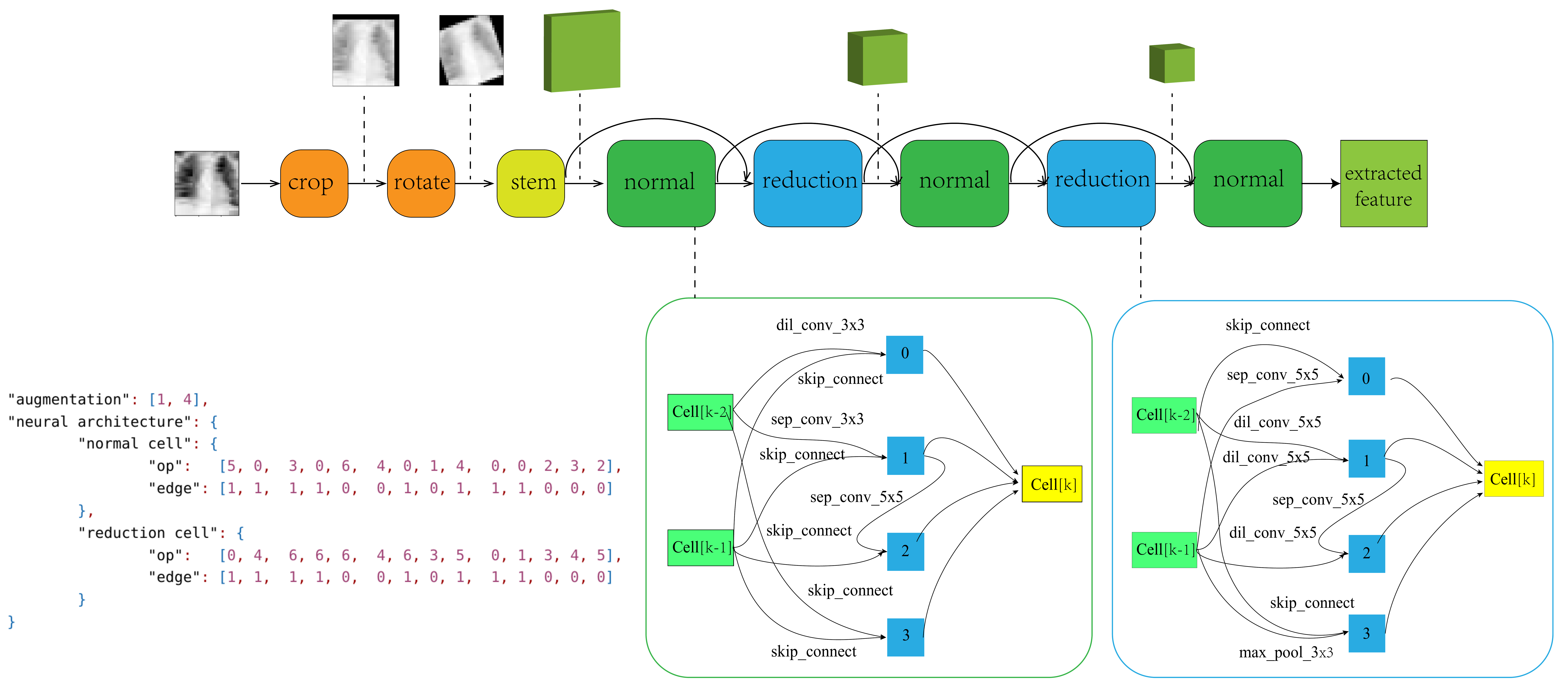}
	\caption{Here is an illustrative example of an individual's encoding. The JavaScript Object Notation (JSON) information located at the left bottom is the individual's encoding. The corresponding network, which contains the augmentation operations and neural architecture, is drawn on the top. In addition, some feature maps in the forward process and the details of the cell structure are indicated through dotted lines.}
	\label{fig_encoding}
\end{figure*}

\subsubsection{The Dark Side of Density Matching}\label{darkside}
The goal of automated data augmentation is defined to find an augmentation policy $T$ with the best performance from a search space $\mathcal{S}_a$ when trained with a model whose parameters are denoted by $\theta$. The optimal augmentation policy optimization problem could be formulated as:
\begin{equation}\label{eq1}
\mathcal{T}^{*} = {argmax}_\mathcal{T}{R(D_{val}, {\theta}^*(\mathcal{T}(D_{train}))))},
\end{equation}
where $R$ is the performance metric function. ${\theta}^*(\mathcal{T}(D_{train}))$ are the weights trained with an augmentation policy $\mathcal{T}$ on $D_{train}$. An augmentation policy contains all the information about data augmentation required during the train process.

To avoid the repeated training procedures for ${\theta}^*(T(D_{train}))$, Density Matching strategy\cite{lim2019fast} was proposed to approximately minimize the distance between $\mathcal{T}(D_{val})$ and $D_{train}$. The objective of the optimization problem in Eq.~(\ref{eq1}) is transformed as:
\begin{equation}\label{eq2}
\mathcal{T}^{*} = {argmax}_{\mathcal{T}}R({\mathcal{T}}(D_{val}),  {\theta}^*(D_{train})).
\end{equation}

Nevertheless, in Density Matching strategy\cite{lim2019fast}, there exists a latent hypothesis. At first, in deep learning, the training process could be formulated as:
\begin{equation}\label{eq3}
{\theta}^{*} = {argmin}_{\theta}\ell(D_{train}),
\end{equation}
where $\ell(\cdot)$ is the loss function and ${\theta}^*(D_{train})$ in Eq.~(\ref{eq2}) is denoted by ${\theta}^{*}$ for simplicity. According to Vapnik-Chervonenkis (VC) Dimension theory\cite{vapnik1999overview, he2019data}, with probability $1 - \delta$, a upper bound of the difference between the expected risk and empirical risk could be formulated as:
\begin{equation}\label{eq4}
\ell(D_{val}, {\theta}^{*}) - \ell(D_{train}, {\theta}^{*}) \leq O(( \frac{|\mathcal{F}|_{VC} - \log{\delta}}{N_{train}})^{\alpha}),
\end{equation}
where $\mathcal{F}$ is hypothesis space of the function $f$ of the model parameterized by $\theta$, $|\mathcal{F}|_{VC}$ is the finite VC-Dimension of $\mathcal{F}$, $N$ is the number of train samples, and $\frac{1}{2} \leq \alpha \leq 1$.

When $\theta$ converges to ${{\theta}}^{*}$, $\ell(D_{train}, {\theta})$ reaches a very small value while $R(D_{train},  {\theta}^*)$ reaches a large one. Without loss of generality, we have: 
\begin{equation}\label{eq5}
R(D_{train},  {\theta}^*) - R(D_{val},  {\theta}^*) \propto O(( \frac{|\mathcal{F}|_{VC} - \log{\delta}}{N_{train}})^{\alpha}).
\end{equation}

We hypothesize that enough great number $N$ of the train samples is given, we have:
\begin{equation}\label{eq6}
R(D_{train},  {\theta}^*) \approx R(D_{val}, {\theta}^*) \geq R(\mathcal{T}(D_{val}), {\theta}^*).
\end{equation}
 
When $R(\mathcal{T}(D_{val}), {\theta}^*)$ is higher, $R(\mathcal{T}(D_{val}),  {\theta}^*)$ is closer to $R(D_{val},  {\theta}^*)$, the corresponding $\mathcal{T}$ could be considered to a more ideal augmentation policy which maintains the density of original distribution. This is how the Density Matching strategy in Eq.~(\ref{eq2}) works.

The Density Matching strategy works well under the condition that enough large training samples are given. However, for small-scale medical image datasets, the in-domain sampling bias demonstrated in Subsubsection~\ref{indomainbias} could cause the inconsistency between $R(D_{train}, {\theta}^*)$ and $R(D_{val}, {\theta}^*)$, which may weaken the efficiency of Density Matching strategy.

\subsubsection{Improving Density Matching by Training with Augmented Data}\label{improving_density_matching}
To solve the disadvantage of Density Matching, a natural solution named \textbf{Augmented Density Matching} is proposed in this paper by increasing the amount of train data through augmentation. Meanwhile, to prevent the search bias, $\theta$ is trained with a prior augmentation policy distribution instead of a fixed one: 
\begin{align}
\begin{split}\label{eq7}
&R(D_{train}, {\theta}^*({\mathcal{T}_{prior}}(D_{train})))) - R(D_{val}, {\theta}^*({\mathcal{T}_{prior}}(D_{train})))) \\
&\propto O(( \frac{|\mathcal{F}|_{VC} - \log{\delta}}{{M}\times{N_{train}}})^{\alpha}),\\
&{\theta}^*({\mathcal{T}_{prior}}(D_{train}) = {argmin}_{\theta}E_{\mathcal{T} \sim \Gamma}({\ell(\mathcal{T}(D_{train}))}),
\end{split}
\end{align}
where $M > 1$ and $\Gamma$ is prior distribution of augmentation policy. In this paper, the above training process is practically implemented by uniformly sampling an augmentation policy from a population of policies.

Therefore, the Augmented Density Matching strategy proposed in this paper could be formulated as:
\begin{equation}\label{eq9}
\mathcal{T}^{*} = {argmax}_{\mathcal{T}}R(\mathcal{T}(D_{val}),  {\theta}^*({\mathcal{T}_{prior}}(D_{train})))).
\end{equation}

\subsection{Unifying the Search on Data Augmentation and Neural Architecture}\label{usaa}
In this subsection, the proposed search framework name USAA, which could search the data augmentation and neural Architecture at the same time, is detailed. To make the search algorithm read practically, the search space is described in Subsubsection~\ref{searchspace} at first. Then, the search framework is detailed in Subsubsection~\ref{searchframework}.

\subsubsection{Search Space} \label{searchspace}
In this subsubsection, the search spaces of augmentation and neural architecture are detailed, respectively.

For the search space of augmentation, the goal is to find an optimal augmentation policy $\mathcal{T}$ consisting of $K$ sub-policies, where each sub-policy $S^{(k)}$ is a sequence of ${L_a}$ different augmentation operations $o_1^{(k)}, o_2^{(k)}, ..., o_{L_a}^{(k)}$, $o_i^{(k)} \in \mathcal{O}_a$, $k=1, 2, ..., K$, $i=1, 2, ..., L_a$, and $\mathcal{O}_a$ denotes the set of all candidate augmentation operations. In this paper, $\mathcal{O}_a$ is constituted by \emph{Indentity} and several most popular augmentation techniques utilized in computer vision tasks, such as \emph{RandomCrop}, \emph{HorizontalFlip}, \emph{VerticalFlip}, \emph{RandomRotate}, \emph{Cutout}\cite{devries2017improved} and \emph{ColorJitter}. \emph{Indentity}, which means no augmentation transform, is added to relax the restraint for searching simple policy. To limit meaningless same operation combinations such as double \emph{HorizontalFlip} and reduce the probability of extreme combinations such as double \emph{Cutout}, the sub-policy is designed as a $L_a$-permutation of $\mathcal{O}_a$ except for consecutive \emph{Indentity}. A normalization is always added after all the augmentation operations. Therefore, the size of the augmentation search space is roughly $(P_{L_a}^{7})^K$, where $P_{L_a}^{7}$ is the number of the $L_a$-permutation of $\mathcal{O}_a$ whose cardinality is 7.

The design of the augmentation search space is inspired by FastAugment\cite{cubuk2020randaugment}, which has demonstrated that it is sufficient to search a single magnitude for all operations. As most augmentation operations in our search space do not have a magnitude, we further abandon the search for magnitude and apply a relatively small magnitude for operations in case of need. Moreover, the probability of operations is also abandoned but implicitly searched by maintaining a population of augmentation sub-policies. This design makes it easy to embed the augmentation search into the EA-based search framework\cite{zhang2021one}.

For the search space of neural architecture, following the settings in \cite{liu2018darts, zhang2021one}, we search a cell-based DARTS search space. The candidate neural architectures are constructed by $L_n$ repeating identity structure, which is referred to as \emph{cell}. Each cell is a directed acyclic graph of seven nodes, with each node is essentially the feature maps. There are two input nodes, one output node, and four intermediate nodes. The two input nodes are the output nodes of the previous two cells, and the output node is the depth-wise concatenation of the intermediate nodes. The first intermediate node is computed based on the two input nodes, while the others are computed based on one or two previous nodes. All the neural operations between the two nodes are chosen from a set $\mathcal{O}_n$ of seven candidate neural operations, where $\mathcal{O}_n =$ \{\emph{skip connection}, \emph{$3 \times 3$ average pooling}, \emph{$3 \times 3$ max pooling}, \emph{$3 \times 3$ and $5 \times 5$ separable convolution}, \emph{$3 \times 3$ and $5 \times 5$ dilated convolution}\}. To get feature maps with a low resolution after the forward computation, there are two types of cells: normal cells and reduction cells. The reduction cells are located at the $1/3$ and $2/3$ depth of the network. The resolutions of feature maps outputted from reduction cells are halved while the channels are doubled. Therefore, the size of the neural architecture search space is roughly $(7^2 + \prod_{n=3}^5({\tbinom{n}{1}\times7 + \tbinom{n}{2}\times7^2}))^2$, where $\tbinom{n}{k}$ is the binomial coefficient.

As medical image datasets always have a wide range of data scales, it is crucial to search for a neural architecture with proper neural capacity. Fortunately, with the DARTS search space, it is easy to adjust the neural capacity by tuning the number of cells $L_n$. Please note that the number of augmentation layers $L_a$ and the number of cells $L_n$ are in a small range and could be searched through a minimal grid search.

\subsubsection{Search Framework}\label{searchframework}
In this subsubsection, the overview of the search framework is outlined at first. Then, some essential steps are detailed in order.

\textbf{Overview of the Search Framework.} The overview of the proposed search algorithm could be seen in Algorithm~\ref{algorithm1}. Firstly, a population of networks is initialized based on a flexible individual encoding strategy (line \ref{line_1}). Then, the population begins to generate and evolve through some generations (lines \ref{line_2}-\ref{line_7}). Finally, the survived population is outputted as the result, which is essentially a population of networks with a fixed neural architecture and a number of sub-policies (line \ref{line_8}).

\textbf{Population Initialization.} In the proposed algorithm, a population of networks are maintained, with each individual contains the information of augmentation sub-policy and neural architecture. Each individual is encoded by two parts: 1) augmentation part, which is essentially a $L_a$-dimensional vector $v^{aug}$, where the $i$-th element of  $v^{aug}(i) \in \{-1, 1, 2, ..., |\mathcal{O}_a|\}$, $|\mathcal{O}_a|$ is the cardinality of $\mathcal{O}_a$. When $v^{aug}(i) >= 1$, it means that operation index of $\mathcal{O}_a$. When $v^{aug}(i) = -1$, it means randomly sampling an operation from $\mathcal{O}_a$. 2) neural encoding part, which has two types of cell encoding parts with same structure for normal cell and reduction cell, respectively. Each cell is encoded by a 14-dimensional $v^{op}$ to denote the neural computational operations and a 14-dimensional $v^{edge}$ to denote the topological connection edges, where the $i$-th element of $v^{op}(i) \in \{-1, 1, 2, ..., |\mathcal{O}_n|\}$, and the $i$-th element of $v^{edge}_i\in\{-1, 0, 1\}$. $v^{edge}_i=0$ and $v^{edge}_i=1$ mean that the $i$th edge of the cell is deactivated and activated respectively. When $v^{op}(i) >= 1$, it means that operation index of $\mathcal{O}_n$. When $v^{op}(i) = -1$, it means randomly sampling an operation from $\mathcal{O}_n$. When $v^{aug}(i) = -1$, we randomly active one or two edges for the corresponding node. To make it more clear, a practical illustration of the individual's encoding and it's corresponding network is shown in Fig.~\ref{fig_encoding}. 

With the help of random code $-1$, the individual encoding is easy to represent a specific network or a group of networks. At the beginning of the search, the population is initialized with one encoding whose augmentation part set as $L_a$ \emph{Indentity} operations to avoid the possible training bias, and all the other parts are filled with random codes. As a result, only one individual encoding is needed to maintained, which greatly simplifies the population representation.

\renewcommand{\algorithmicrequire}{\textbf{Input:}} 
\renewcommand{\algorithmicensure}{\textbf{Output:}} 
\begin{algorithm}[H] 
	\caption{Unified Search Framework} 
	\label{algorithm1} 
	\begin{algorithmic}[1] 
		\Require 
		The number of generations $l$;
		The train and validation set $D_{train}$, $D_{val}$, respectively;
		The number of outputted sub-policies $K$.
		\Ensure 			
		$P$;
		\State initial $P$ \Comment{population initialization} \label{line_1}
		
		\For{$i=1$ to $l$}  \label{line_2} 
		
		\State $\theta$ = Train($P$, $D_{train}$) \Comment{supernet training stage} \label{line_3}
		
		\State $P$ = Generation($P$)
		\Comment{generation stage}
		\State $Fitness_{P}$ = Evaluate($\theta$, $P$, $D_{val}$) \Comment{fitness evaluation for $P$}
		
		\State $P$ = Evolution($P, Fitness_{P}$) \Comment{environmental selection and evolution} \label{line_6}
		
		\EndFor  \Comment{$P$ will be a population of networks with a fix neural architecture and a number of sub-policies after the last evolution} \label{line_7}
		
		\State select the best $K$ individuals from $P$ \label{line_8} 
	\end{algorithmic} 
\end{algorithm}

\begin{figure}
	\includegraphics[width=0.49\textwidth]{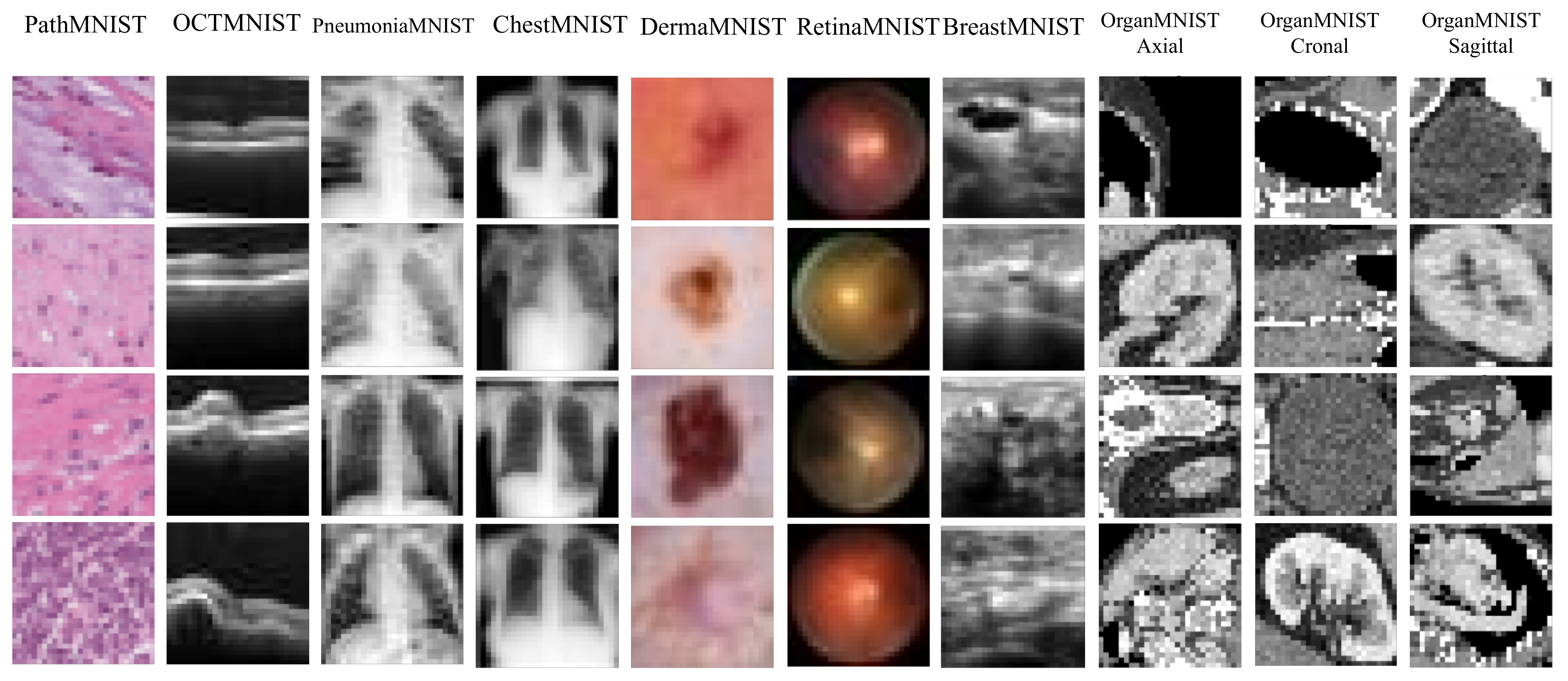}
	\caption{Examples from MedMNIST. Each column has four examples with their corresponding sub-dataset name shown on the top.}
	\label{fig_datasetshow}
\end{figure}

\textbf{Training Supernet.} The goal of this step is to warm up the weights $\theta$ of supernet, which are expected to judge candidate networks fairly. In the supernet training stage, a number of train epochs are performed to warm up the sharing weights. In the meantime, following \cite{bender2018understanding, guo2019single, li2020improving, zhang2021one}, a path dropout strategy is always utilized to make the sharing weights more predictive. In this paper, this strategy is implemented by randomly sample a candidate network from the maintained population, which could be formulated as below:
\begin{equation}\label{eq10}
{\theta}^* = {argmin}_{\theta}E_{\mathcal{N} \sim \mathcal{P}}({\ell(D_{train}, \mathcal{N})}),
\end{equation}
where $\mathcal{N}$ is the sampled candidate network, which contains the information of both augmentation sub-policy and neural architecture. $\mathcal{P}$ is the distribution of maintained population. Referring to Eq.~(\ref{eq7}) and (\ref{eq10}), it is obvious to find the analogy between them. Despite the different purposes, the process of supernet training could be formulated in a unified way. 

\textbf{Generation.} In the generation stage, different individuals are generated in the population. Following \cite{zhang2021one}, in each round of generation, only one encoding part of the parent individual is generated while the others are fixed. What is more, the population generates in a higher-to-lower order: first generate the neural computational operations in higher layers(closer to the output node), then generate those in lower layers, next generate the edges and finally generate the augmentation operations in the same order. As the random codes keep the population diversity, a parent individual could generate its children without crossover. The total number of generations $l$ could be calculated by $L_a + (14 + 3)\times2$, where 14 denotes the number of generations for neural computational operations, 3 denotes that for the edge search for the last three intermediate nodes, and 2 denotes a double calculation for normal and reduction cell.

\renewcommand\tabcolsep{6.0pt} 
\begin{table}\label{table_datasetinfo}
	\centering 
	\caption{Some specific information about MedMNIST} 
	\begin{threeparttable}
		\begin{tabular}{@{}lllr@{}}
			\toprule
			Sub-dataset & task type & CB ratio\tnote{+} & data scale \\ 
			\midrule
			PathMNIST  & multi-class               & 1.63 & 107180 \\
			ChestMNIST & multi-label & 8.81      & 112100 \\ 
			DermaMNIST & multi-class               & 58.66     & 10015  \\  
			OCTMNIST   & multi-class               & 5.94              & 109309 \\
			PneumoniaMNIST & binary-class          & 2.88      & 5856   \\ 
			
			RetinaMNIST & ordinal regression       & 7.36    & 1600  \\ 
			BreastMNIST & binary-class             & 2.71 & 780   \\ 
			OrganMNIST\_Axial & multi-class        & 4.54     & 58850 \\ 
			OrganMNIST\_Coronal & multi-class      & 4.98     & 23660 \\ 
			OrganMNIST\_Sagittal & multi-class     & 5.63     & 25221 \\ 
			\bottomrule
		\end{tabular}
		
		\begin{tablenotes}
			\footnotesize
			\scriptsize{
				\item[+] The class imbalance ratio is calculated through dividing the number of majority class samples by the number of minority class samples.
			}
		\end{tablenotes}
	\end{threeparttable}
\end{table}

\renewcommand\tabcolsep{6.0pt} 
\begin{table*}
	\centering 
	\caption{Overall results on MedMNIST}\label{table_overall} 
	\begin{tabular}{@{}ccccccccccccccc@{}}
		\toprule
		\multirow{2}{*}{Methods} & 
		\multicolumn{2}{l}{PathMNIST} & 
		\multicolumn{2}{l}{ChestMNIST} & 
		\multicolumn{2}{l}{DermaMNIST} & 
		\multicolumn{2}{l}{OCTMNIST} & 
		\multicolumn{2}{l}{\fontsize{24pt}{24pt}{PneumoniaMNIST}} \\ 
		& AUC & ACC & AUC & ACC & AUC & ACC & AUC & ACC & AUC & ACC \\
		
		\midrule
		ResNet-18 (28)  & 0.979 & 0.860  & 0.706 & 0.947  & 0.911 & 0.750  & 0.951 & 0.758  & 0.795 & 0.843\\
		ResNet-18 (224)  & 0.978 & 0.860 & 0.713 & 0.948 & 0.896 & 0.727 & 0.960 & 0.752 & 0.814 & 0.861\\
		ResNet-50 (28)  & 0.969 & 0.846 & 0.692 & 0.947 & 0.899 & 0.727 & 0.939 & 0.745 & 0.813 & 0.857\\
		ResNet-50 (224)  & 0.978 & 0.848 & 0.706 & 0.947 & 0.895 & 0.719 & 0.951 & 0.750 & 0.865 & 0.896\\
		auto-sklearn\cite{feurer5872efficient}  & 0.500 & 0.186 & 0.647 & 0.642 & 0.906 & 0.734 & 0.883 & 0.595 & 0.824 & 0.865\\
		AutoKeras\cite{jin2019auto}  & 0.979 & 0.864 & 0.715 & 0.939 & 0.921 & 0.756 & 0.956 & 0.736 & 0.897 & 0.918\\
		Google AutoML Vision  & 0.982 & 0.812 & 0.718 & 0.947 & \textbf{0.925} & 0.761 & 0.965 & 0.732 & \textbf{0.993} & 0.941\\
		USAA  & \textbf{0.989} & \textbf{0.931} & \textbf{0.798} & \textbf{0.948} & 0.912 & \textbf{0.768} & \textbf{0.967}	& \textbf{0.833} & 0.985 & \textbf{0.942} \\

		\midrule
		\multirow{2}{*}{Methods} & 
		\multicolumn{2}{l}{RetinaMNIST} & 
		\multicolumn{2}{l}{BreastMNIST} & 
		\multicolumn{2}{l}{OrganMNIST\_A} & 
		\multicolumn{2}{l}{OrganMNIST\_C} & 
		\multicolumn{2}{l}{OrganMNIST\_S} \\ 
		& AUC & ACC & AUC & ACC & AUC & ACC & AUC & ACC & AUC & ACC \\

		\midrule		
		ResNet-18 (28)  & 0.723 & 0.530 & 0.821 & 0.859 & 0.995 & 0.921 & 0.990 & 0.889  & 0.967  & 0.762 \\
		ResNet-18 (224) & 0.721 & 0.543 & 0.857 & 0.878 & 0.997 & 0.931
		& 0.991 & 0.907 & 0.974 & 0.777\\
		ResNet-50 (28)  & 0.728 & 0.518 & 0.839 & 0.853 & 0.995 & 0.916 & 0.990 & 0.893 & 0.968 & 0.746\\
		ResNet-50 (224)  & 0.717 & \textbf{0.555} & 0.818 & 0.833 & 0.997 & 0.931 & \textbf{0.992}
		& 0.898 & 0.970 & 0.770\\
		auto-sklearn\cite{feurer5872efficient}  & 0.694 & 0.525 & 0.673 & 0.808 & 0.797 & 0.563 & 0.898 & 0.676 & 0.855 & 0.601\\
		AutoKeras\cite{jin2019auto}  & 0.655 & 0.420 & 0.646 & 0.801 & 0.996 & 0.929
		& \textbf{0.992} & 0.915 & 0.972 & 0.803\\
		Google AutoML Vision  & 0.759 & 0.523 & \textbf{0.932} & 0.865 & 0.988 & 0.816 & 0.986 & 0.862 & 0.964 & 0.707\\
		USAA  & \textbf{0.788} & 0.55 & 0.92 & \textbf{0.91} & \textbf{0.997} & \textbf{0.959} & 0.991 & \textbf{0.926} & \textbf{0.981} & \textbf{0.838} \\
		
		\bottomrule
	\end{tabular}
\end{table*}

\textbf{Fitness Evaluation and Evolution.} After generation, the fitness of the generated individual is evaluated as the metrics to decide which individual to survive or to be weeded out. Due to AUC is the most popular metric for class imbalance problems\cite{zhao2011online}, which is a natural feature of most medical image datasets, AUC is selected as the primary evaluation standardization. At the same time, accuracy (ACC) serves as an additional metric. As a result, it could be formulated as a multi-objective optimization problem:
\begin{equation}\label{eq11}
\begin{cases}
f1: {maximize}_{\mathcal{N} \sim \mathcal{P}}AUC(\mathcal{N}, {\theta}^*) \\
f2: {maximize}_{\mathcal{N} \sim \mathcal{P}}ACC(\mathcal{N}, {\theta}^*)
\end{cases}, 
\end{equation}
where $AUC(\mathcal{N}, {\theta}^*)$ and $ACC(\mathcal{N}, {\theta}^*)$ are the calculated based on validation set and the weights got in previous supernet training stage.

After the fitness evaluation, following \cite{zhang2021one}, a multi-objective evolutionary algorithm NSGA-II\cite{deb2002fast} is utilized to optimize the population. As the goal of the proposed algorithm is to find an optimal network, which is constituted by one neural architecture and $K$ augmentation sub-policies, a dynamical population size after evolution is adjusted to survive the best individual after generating neural architecture part and $K$ best individuals after generating the augmentation part.

\section{Experiments And Results}\label{experiment}
To evaluate the efficiency of the proposed algorithm, a series of experiments have been performed. Firstly, the experimental design is described in subsection \ref{experiment_design}. Then, the experiment results and the ablation studies are described in subsection \ref{results} and \ref{ablation}, respectively. Finally, some in-depth analyses are discussed in subsection \ref{indepth}.

\subsection{Experimental Design}\label{experiment_design}
\subsubsection{Benchmark Dataset}
The MedMNIST\cite{yang2021medmnist} is chosen as the benchmark in the experiments. MedMNIST could be considered the first pioneering benchmark dataset specially constructed to compare AutoML methods in medical image analysis. It consists of 10 sub-datasets from different medical research areas.
Some examples from MedMNIST are shown in Fig.~\ref{fig_datasetshow} for reference, and some specific information is shown in Table~\ref{table_datasetinfo}. 

The reasons for choosing MedMNIST are: 1) it covers primary medical image modalities, such as histology slides, X-ray images, dermatoscopic images, optical coherence tomography (OCT) images, fundus images, ultrasound images, and computed tomography (CT) images. 2) it has a diverse data scale range from about 500 to 100000 and tasks consisting of binary-class, multi-class, ordinal regression, and multi-label. 3) all the sub-datasets are pre-processed to a resolution of $28\times28$, making it easy to be utilized with no background knowledge.

\subsubsection{Peer Competitors}
In order to demonstrate the efficiency of the proposed algorithm, some representative and state-of-the-art algorithms are chosen as the peer competitors for comparison. The chosen methods could be summarized into three categories.

The first category is two versions of ResNet\cite{he2016deep} with 18 layers and 50 layers, named ResNet-18 and ResNet-50, respectively, which are used as the baseline to compare the state-of-the-art manually designed convolutional neural network (CNN). ResNet is proper to handle the images with low resolutions, and two different versions could help to exam the scale-diverse sub-datasets with two choices of model capacity. What is more, two resolution sizes of the input images, i.e., $28\times28$ and $224\times224$, are tested to further evaluate ResNet's performance.

The second category refers to the most representative AutoML tools, such as Auto-sklearn\cite{feurer5872efficient}, AutoKeras\cite{jin2019auto} and Google AutoML Vision. Auto-sklearn could be considered as the representative of open-source AutoML tool for conventional machine learning, while AutoKeras is the representative for deep learning. Besides, Google AutoML Vision is a representative commercial AutoML tool for deep learning. As all the above tools have integrated many AuotML techniques such as feature engineering, model selection, and algorithm selection, they are appropriate for comparing overall results. 

The third category contains the recent state-of-the-art NAS and auto-augmentation methods, such as HOPNAS\cite{zhang2021one}, FastAugment\cite{lim2019fast}, FasterAugment\cite{hataya2020faster}. HOPNAS is introduced as the baseline of NAS to demonstrate the efficiency of unifying the search on data augmentation and neural architecture. Furthermore, the other auto-augmentation methods are introduced to demonstrate the proposed Augmented Density Match strategy. For a fair comparison, all the auto-augmentation methods are embedded with the same neural architectures searched by the proposed method.  
\begin{figure}
	\centering
	\includegraphics[width=0.5\textwidth]{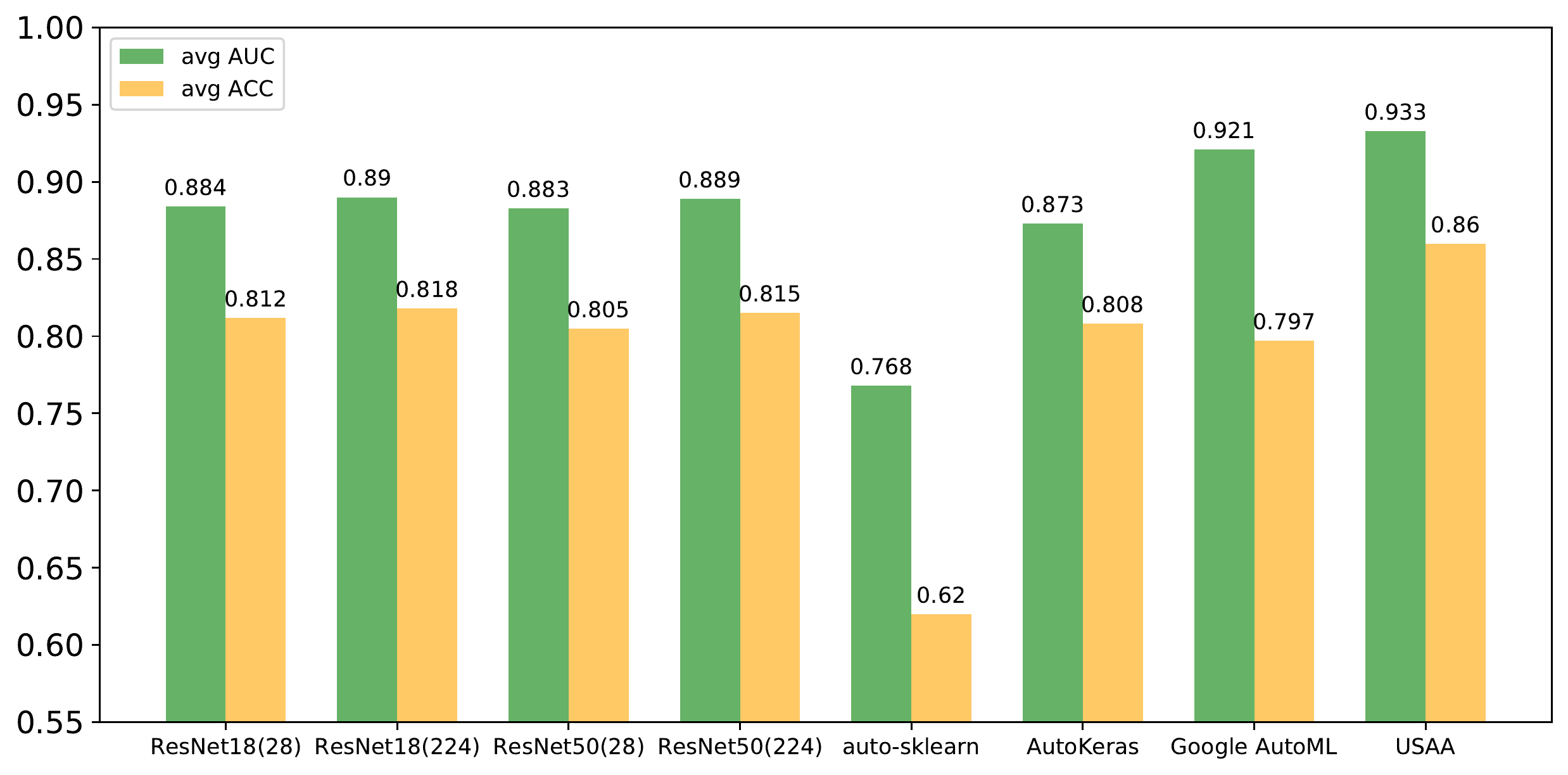}
	\caption{Comparison average AUC/ACC on MedMNIST.}
	\label{fig_average}
\end{figure}

\subsubsection{Parameter Settings} 
For the parameter settings about search space, the number of sub-policies of one augmentation policy is set as 10 to follow the previous works\cite{lim2019fast, hataya2020faster}. To constrain the augmented data close to the origin, the range of the augmentation operation number $L_a$ is set from 1 to 3. To properly adjust the searched model's capacity, the number of the cells $L_n$ is searched in a range from 1 to 12.

For the proposed search method, during the supernet training stage in Algorithm~\ref{algorithm1}, the number of training epochs is set as 20 at the first generation, while that is set as 5 for the others. For the generation stage, the size of the maintained population $N_p$ is set as 7 when searching neural architecture part following \cite{zhang2021one}, while that is set as 30 when searching augmentation part. During the whole search stage, following \cite{yang2021medmnist}, the batch size is set to 128. SGD optimizer with a momentum of 0.9 and an L2 regularization of 3e-4 is utilized with an initial learning rate of 0.025. A cosine schedule is utilized for learning rate decay. A max search trials of 10 is set for the grid search on $L_a$ and  $L_n$.

When evaluating the searched network, following \cite{yang2021medmnist}, all the models are trained of 100 epochs, and the other training parameters are set the same as in the search stage. The model with the highest AUC on validation set is selected for test. In addition, if two models' AUC are quite similar within a distance of 0.001, the model with better generalization performance, i.e., more minor difference of validation loss minus train loss, is chosen as the output. Please note that the test set has never participated in the search or the model selecting process.

\subsection{Results}\label{results}
\subsubsection{Overall Results}\label{overall_results}
The overall comparison results on MedMNIST are shown in Table~\ref{table_overall}, and the comparison average AUC and ACC are shown in Fig~\ref{fig_average}. According to Table~\ref{table_overall} and Fig~\ref{fig_average}, the four groups of ResNet experiments  yield different performance on the ten sub-datasets but similar overall results. It seems hard for a manually designed network to make a trade-off between all the sub-datasets by simply tuning the model capacity or input resolution size, which is consistent with the No Free Lunch Theorem\cite{wolpert1997no}.  Therefore, it is significant to search specific augmentation policies and neural architectures for different medical image datasets.

Auto-sklearn got the weakest performance, which may be interpreted that the conventional AutoML tool based on statistical machine learning is not competent for handling complex medical image analysis problems.  Neither AutoKeras nor Google AutoML Vision can completely suppress manually designed ResNet on all the sub-datasets. However, the proposed methods outperformed ResNet on all the metrics except for two terms. The average AUC of the proposed method is $1.2\%$ higher than Google AutoML Vision and $6\%$ higher than AutoKeras, while the average ACC of the proposed method is $6.3\%$ higher than Google AutoML Vision and $5.2\%$ higher than AutoKeras. In summary, the overall results of the proposed methods outperformed all versions of ResNet, auto-sklearn\cite{feurer5872efficient}, AutoKeras\cite{jin2019auto} and Google AutoML Vision.

\renewcommand\tabcolsep{2.0pt} 
\begin{table}
	\scriptsize
	\centering 
	\caption{Comparison on Different Augmentation Strategies}\label{table_augment} 
	\begin{tabular}{@{}llccccc@{}}
		\toprule
		\multicolumn{2}{l}{Methods} & HOPNAS\cite{zhang2021one} & Rand &
		\tabincell{c}{FastAug\cite{lim2019fast}}
		& FasterAug\cite{hataya2020faster} & USAA \\ 
		\midrule
		
		\multirow{2}{*}{PathMNIST}  & AUC & 0.987 & 0.988 & 0.985 & 0.984 & \textbf{0.989} \\ 
		& ACC & 0.912 & 0.922 & 0.883 & 0.892 & \textbf{0.931} \\ 
		\midrule
		\multirow{2}{*}{ChestMNIST} & AUC & 0.763 & 0.782 & 0.782 & 0.757 & \textbf{0.798} \\ 
		& ACC & 0.947 & 0.948 & 0.948 & 0.948 & \textbf{0.948} \\ 
		\midrule
		\multirow{2}{*}{DermaMNIST} & AUC & 0.899 & 0.893 & 0.902 & 0.89  & \textbf{0.906} \\ 
		& ACC & 0.759 & 0.751 & \textbf{0.767} & 0.733 & 0.763 \\ 
		\midrule
		\multirow{2}{*}{OCTMNIST}   & AUC & 0.948 & 0.951 & 0.958 & 0.955 & \textbf{0.967} \\ 
		& ACC & 0.761 & 0.797 & 0.777 & 0.787 & \textbf{0.833} \\ 
		\midrule
		\multirow{2}{*}{PneumoniaMNIST}& AUC & 0.971 & 0.964 & 0.963 & 0.966 & \textbf{0.982}\\ 
		& ACC & 0.852 & 0.856 & 0.859 & 0.889 & \textbf{0.896}\\ 
		\midrule
		\multirow{2}{*}{RetinaMNIST} & AUC & 0.77 & \textbf{0.793} & 0.758 & 0.767 & 0.792 \\ 
		& ACC & 0.523 & 0.48 & 0.49  & 0.455 & \textbf{0.523} \\ 
		\midrule
		\multirow{2}{*}{BreastMNIST} & AUC & 0.907 & 0.878 & 0.912 & \textbf{0.912} & 0.907 \\ 
		& ACC & 0.853 & \textbf{0.859} & 0.853 & 0.897 & 0.853 \\ 
		\midrule
		\multirow{2}{*}{OrganMNIST\_A} & AUC & 0.995 & 0.989 & 0.996 & 0.995 & \textbf{0.997}\\ 
		& ACC & 0.937 & 0.937 & 0.954 & 0.947 & \textbf{0.959}\\ 
		\midrule
		\multirow{2}{*}{OrganMNIST\_C} & AUC & 0.988 & 0.985 & 0.99  & 0.989 & \textbf{0.991}\\ 
		& ACC & 0.911 & 0.914 & 0.925 & 0.917 & \textbf{0.926}\\ 
		\midrule
		\multirow{2}{*}{OrganMNIST\_S} & AUC & 0.975 & 0.968 & 0.975 & 0.975 & \textbf{0.977}\\ 
		& ACC & 0.803 & 0.814 & 0.82  & 0.809 & \textbf{0.825}\\ 
		\midrule
		\multirow{2}{*}{Average} & AUC & 0.92 & 0.919 & 0.922 & 0.919 & \textbf{0.931}\\ 
		& ACC & 0.826 & 0.828 & 0.828 & 0.827 & \textbf{0.846}\\ 
		
		\bottomrule
	\end{tabular}
\end{table}

\subsubsection{The Efficiency of Augmented Density Matching}\label{augment_results}
To demonstrate the efficiency of the proposed augmentation search strategy, i.e., Augmented Density Matching, different state-of-the-art auto-augmentation 2 methods have been experimented for comparison. HOPNAS\cite{zhang2021one} is introduced as the baseline, which only searches the neural architecture. FastAugment\cite{lim2019fast} and FasterAugment\cite{hataya2020faster} are introduced as state-of-the-art peer competitors to demonstrate the proposed augmentation search strategy. Random search is also introduced to exam the performance difference between the proposed method and random augmentation policies. To promise comparison fairness, all the methods were experimented with the same group of neural architectures and hyper-parameters, e.g., the number of augmentation sub-policies and operations.

The detailed comparison results are shown in Table~\ref{table_augment}. According to Table~\ref{table_augment}, it could be found that there was a significant gap between the performance of HOPNAS and the proposed unified search framework, i.e., USAA. With a few more generation rounds and little search time for searching augmentation policy, the proposed method could achieve $1.1\%$ higher average AUC and $2\%$ higher average ACC than HOPNAS, which demonstrates that unifying the search on data augmentation and neural architecture is profitable. For auto-augmentation methods, FasterAugment yields the worst performance, quite close to HOPNAS and Random Search. This phenomenon is in line with expectations because FasterAugment utilized the train set to optimize the weights of augmentation operation. However, there is always a large gap between the distributions of train set and validation set for medical images, which leads to a local optimal. The performance of FastAugment is slightly better, which may benefit from the Density Matching strategy. The proposed method achieves a much better performance than FastAugment, which demonstrated the efficiency of Augmented Density Matching.

\renewcommand\tabcolsep{7.0pt} 
\begin{table}
	\centering 
	\caption{Comparison on Different Augmentation Search Space}\label{table_space} 
	\begin{tabular}{@{}lcccc@{}}
		\toprule
		\multirow{2}{*}{Methods} & \multicolumn{2}{c}{Large Search Space} & 
		\multicolumn{2}{c}{Small Search Space} \\ 
		& AUC & ACC & AUC & ACC \\
		
		\midrule
		
		PathMNIST  	    & 0.983& 0.869 & \textbf{0.989} & \textbf{0.931}  \\ 
		ChestMNIST  	& 0.79 & 0.948 & \textbf{0.798}  & 0.948 \\ 
		DermaMNIST  	& 0.859& 0.731 & \textbf{0.906} & 0.763 \\ 
		OCTMNIST        & 0.952& 0.804 & \textbf{0.967} & \textbf{0.833}  \\ 
		PneumoniaMNIST  & 0.959& 0.761 & \textbf{0.982} & \textbf{0.896}  \\ 
		
		RetinaMNIST     & \textbf{0.799} & 0.42  & 0.792& \textbf{0.523}  \\ 
		BreastMNIST     & 0.904& \textbf{0.865}  & \textbf{0.907} & 0.853 \\ 
		OrganMNIST\_A   & 0.996& 0.943 & \textbf{0.997} & \textbf{0.959}  \\ 
		OrganMNIST\_C   & 0.989& 0.919 & \textbf{0.991} & \textbf{0.926}  \\ 
		OrganMNIST\_S   & 0.975& 0.813 & \textbf{0.977} & \textbf{0.825}  \\ 

		\midrule
		Average  & 0.921 & 0.807 & \textbf{0.931} & \textbf{0.846} \\ 

		\bottomrule
	\end{tabular}
\end{table}

\subsection{Ablation Study}\label{ablation}
In this subsection, the impact of the augmentation search space and the sub-policy number are experimented and discussed.

\subsubsection{Augmentation Search Space}
As the search space is critical for searching optimal augmentation, two search spaces are evaluated to investigate the impact of the augmentation search space. The first evaluated search space is the one used in the proposed method and introduced in Subsubsection~\ref{searchspace}, while the second search space, following FasterAugment\cite{hataya2020faster}, is built by adding another 14 candidate augmentation operations based on the first one. The additional augmentation operations are \emph{Posterize, ShearX, ShearY, TranslateY, TranslateY, Invert, Solarize, Contrast, Saturate, Brightness, Sharpness, AutoContrast and Equalize}. For simplicity, the first search space utilized in the proposed methods is referred to as \textbf{Small Search Space} while the second one is referred to as \textbf{Large Search Space} in this paper. Therefore, the cardinality of Small Search Space is 7 while that of Large Search Space is 21.

The comparison results of the two search spaces are shown in Table~\ref{table_space}. It could be found that the performance of Small Search Space is better than that of Large Search Space, although the former search space is larger. It could be interpreted that designing a compacter and proper augmentation search space makes the search less possible to reach local optimal. 

\subsubsection{The Number of Augmentation Sub-policies}
\begin{figure}
	\centering
	\includegraphics[width=0.45\textwidth]{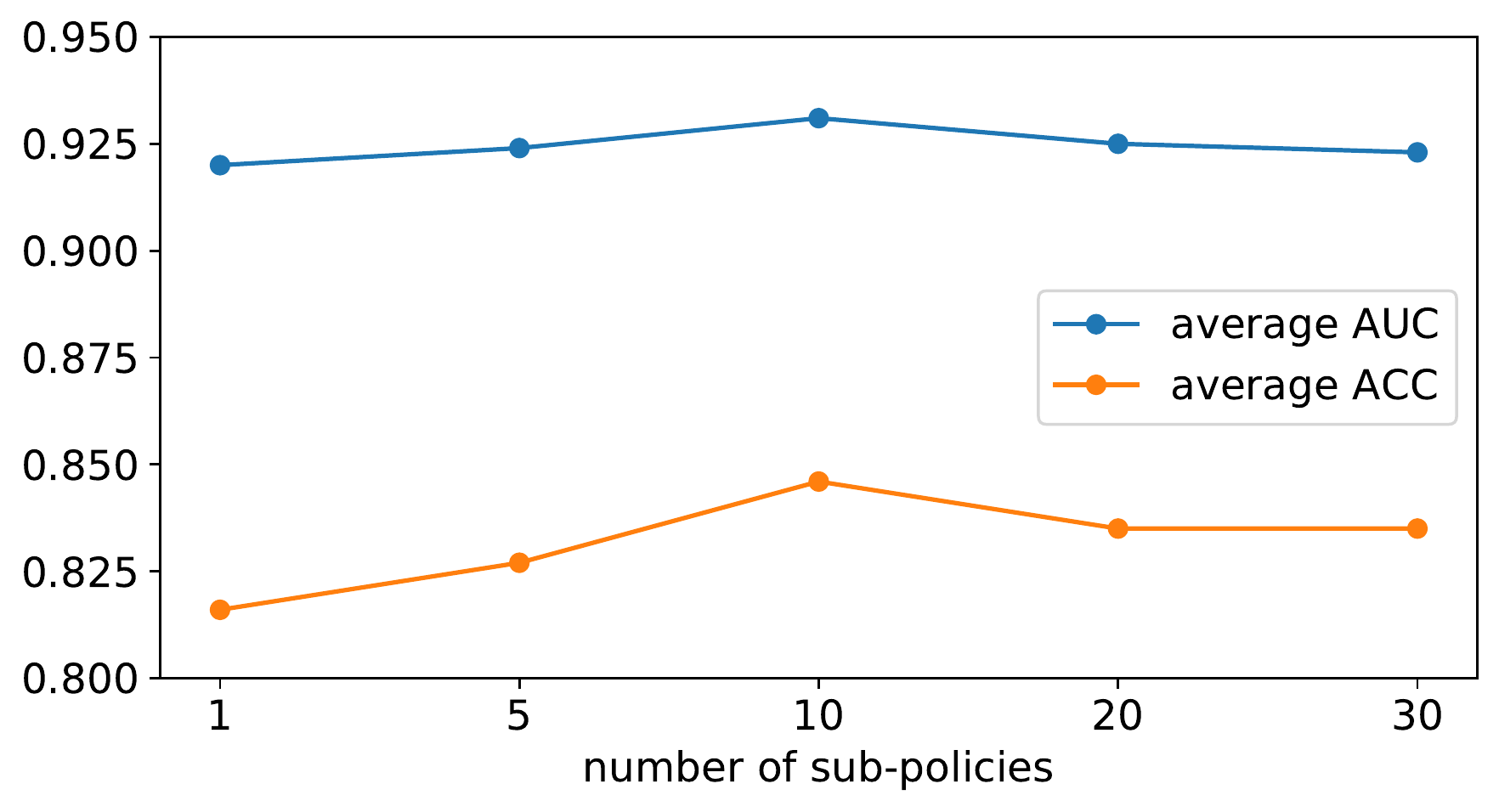}
	\caption{Comparison Results on Different Number of Sub-policies.}
	\label{fig_subpolicynum}
\end{figure}

The number of augmentation sub-policies in one augmentation policy is also investigated in this paper. The average AUC/ACC of different sub-policies numbers are shown in Fig~\ref{fig_subpolicynum}. It could be found that the number of sub-policies is critical to the performance of auto-augmentation. It got the worst AUC and ACC when the number of sub-policies is set as 1. The average performance increase with the number of sub-policies until reaching about 10. A diverse, stochastic mix of augmentation sub-policies could significantly improve the performance, which is consistent with what was found in \cite{cubuk2019autoaugment}. A reasonably large number of sub-policies reduce the probability of local optima and improve the diversity of augmented data.

\subsection{Indepth Analysis}\label{indepth}
In this subsection, the neural capacity and the number of augmentation operations in one sub-policy of searched optimal networks are investigated, and some interesting results are found and discussed. 

It is widely known that there exists an optimal model capacity for a known dataset in deep learning\cite{goodfellow2016deep}. If a model's capacity is larger than optimal model capacity, the training will go into the overfitting regime. Otherwise, the training would go into the underfitting regime. If the range of optimal model capacity could be approximately estimated, the search efficiency would be significantly improved by constraining the search space. However, the optimal capacity is too tricky to estimate but widely thought proportional to the number of training samples. To investigate their relation, we use the cell number of the searched networks to estimate the optimal capacity roughly. Moreover, the augmentation operation number of sub-policy is also investigated. 

\begin{figure}
	\centering
	\includegraphics[width=0.45\textwidth]{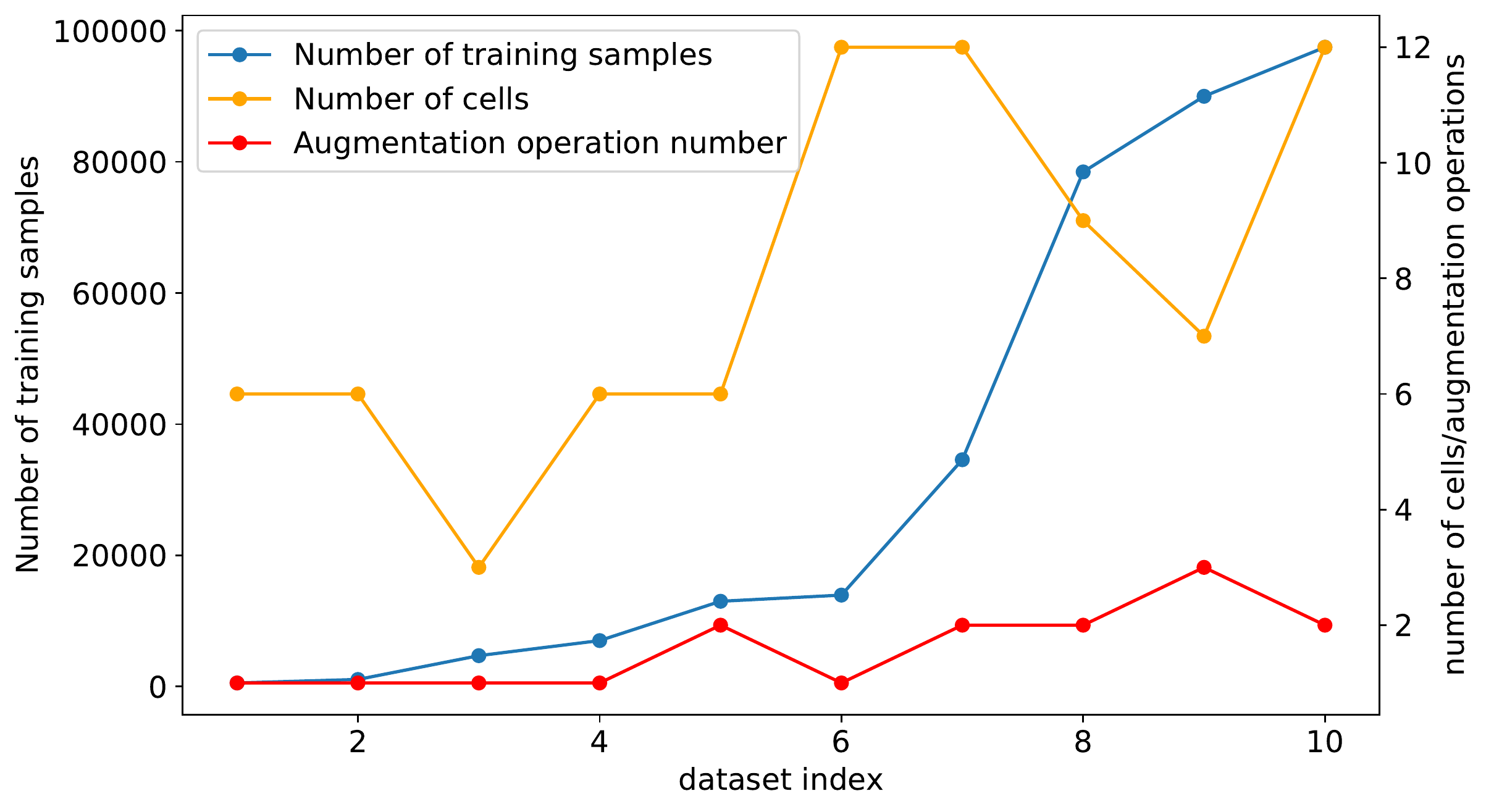}
	\caption{The number of training samples, the cell number, and the number of augmentation operations of the searched networks on the 10 sub-datasets.}
	\label{fig_capacity}
\end{figure}

The number of training samples, the cell number, and the augmentation operation number of the searched networks on the 10 sub-datasets are shown in Fig~\ref{fig_capacity}. It could be found that the cell number of the searched networks on the top 5 small sub-datasets are all no greater than 6, while that on the top 5 large sub-datasets are all no less than 6. The cell number is roughly proportional to the data scale of sub-datasets, which is consistent with the conventional view about optimal neural capacity\cite{goodfellow2016deep}. For the augmentation operation number of searched networks, most small-scale sub-datasets prefer only one augmentation operation, while the large-scale sub-datasets prefer a small larger number of augmentation operations. Except for optimal neural capacity, there may exist an \textbf{optimal augmentation capacity} for a dataset in deep learning.

The above analysis reveals the insight that the data scale should be under consideration when designing AutoML algorithms for diverse-scale medical image datasets. Limiting the search space for both neural architecture and augmentation policy seems like a simple but practical way to improve the efficiency of AutoML methods. 

\section{Conclusion}\label{conclusion}
In this paper, we theoretically and empirically demonstrated the inconsistency between the train and validation set of small-scale medical image datasets, referred to as in-domain sampling bias. An improved augmentation search strategy named Augmented Density Matching is proposed to address the inefficiency of Density Matching caused by in-domain sampling bias. Further, an efficient AutoML algorithm named USAA is proposed by unifying the search on data augmentation and neural architecture. The experiment results have shown that both the proposed Augmented Density Matching and USAA outperformed previous state-of-the-art methods. Moreover, through in-depth analysis we found that, besides optimal neural capacity, there was an optimal augmentation capacity for a dataset in deep learning. Constraining the search space of augmentation and neural capacity in terms of data scale should benefit future works.


%

%


\ifCLASSOPTIONcaptionsoff
  \newpage
\fi




\bibliographystyle{IEEEtran}
\bibliography{IEEEabrv} 

\end{document}